# EVALUATING AI VOCATIONAL SKILLS THROUGH PROFESSIONAL TESTING

David Noever and Matt Ciolino
PeopleTec, Inc., Huntsville, AL, USA
David.noever@peopletec.com    matt.ciolino@peopletec.com

*ABSTRACT*

*Using a novel professional certification survey, the study focuses on assessing the vocational skills of two highly cited AI models, GPT-3 and Turbo-GPT3.5. The approach emphasizes the importance of practical readiness over academic performance by examining the models' performances on a benchmark dataset consisting of 1149 professional certifications. This study also includes a comparison with human test scores, providing perspective on the potential of AI models to match or even surpass human performance in a wide range of professional certifications. GPT-3, even without any fine-tuning or exam preparation, managed to achieve a passing score (over 70% correct) on 39% of the professional certifications. It showcased proficiency in computer-related fields, including cloud and virtualization, business analytics, cybersecurity, network setup and repair, and data analytics. Turbo-GPT3.5, on the other hand, scored a perfect 100% on the highly regarded Offensive Security Certified Professional (OSCP) exam. This model also demonstrated competency in diverse professional fields, such as nursing, licensed counseling, pharmacy, and teaching. In fact, without any prior preparation, it passed the Financial Industry Regulatory Authority (FINRA) Series 6 exam with a grade of 70%. Turbo-GPT3.5 exhibited strong performance on customer service tasks, indicating potential use cases in enhancing chatbots for call centers and routine advice services. Both models also scored well on sensory and experience-based tests outside a machine's traditional roles, including wine sommelier, beer tasting, emotional quotient, and body language reading. The study found that OpenAI's model improvement from Babbage to Turbo led to a 60% better performance on the grading scale within a few years. This progress indicates that addressing the current model's limitations could yield a high-performing AI capable of passing even the most rigorous professional certifications.*

## 1. INTRODUCTION

The increasing attention towards assessing the advancements of large language models (LLMs) using professional exams has opened up new avenues in AI research [1-47]. These exams span a diverse range of disciplines, including but not limited to medicine [6-7,15,17,20,23-25,27-28,30-31,36], law [13], physics [43], mathematics [16,41], engineering [34], software development [14], psychology [32], IQ [39], finance [37], language translation [8,10-12,21,29,44], and general problem-solving tasks [1-2,9,18,26,35,38,42].

In response to the growing capabilities of LLMs, a new research area has emerged with a focus on designing exam formats that prevent students from utilizing LLMs during testing [40]. The initial Turing test [48-49], which evaluated if a machine could convince a human panel that they were conversing with a human and not a machine, laid the groundwork for subsequent evaluation methods. More recent approaches have evaluated LLMs based on grammatical correctness, and in the case of translation tasks, the Bilingual Evaluation Understudy (BLEU) metric has been used for performance comparison against human experts and other machines [8,10-12,21,29,44].

The remarkable strides in language understanding made by models like ChatGPT and other transformers since late 2022 [3-4] have shifted the focus towards the current generation of LLMs. Their impressive ability to pass professional certification exams in complex and critical fields, such as legal advisory [13] or internal medicine [6], has piqued the interest of researchers. Some early prototype systems have even showcased potential applications in sensitive areas like suicide prevention and mental health counseling, provided certain safeguards are in place [50].

Recognizing the critical importance of understanding the training, refining, and deployment process of these models, OpenAI [4] has published its "eval corpus" [51]. This corpus primarily provides a detailed description of the internal stages involved in determining a model's safety for deployment.

This study is inspired by a thought-provoking hypothetical scenario: What if ChatGPT's resume was presented to an employer during a job interview? Would it be feasible to consider the AI model as a reliable personal assistant and a potential new hire? [22] This question gains significance when considering the model's previous achievements in challenging examinations [4].

ChatGPT (GPT 4.0) has passed not only medical [6] and legal bar exams [14] but also secured certification as a competitive programmer on Code Forces [4]. As a math and biology Olympiad winner, GPT4 has successfully cleared all Advanced Placement exams and secured a position in the top 5% of college admission tests (Scholastic Aptitude Test) [4]. In such a scenario, it's not far-fetched to consider that a machine with such credentials could easily qualify for undergraduate and graduate programs [4,42].

Given these accomplishments, could the machine demonstrate equivalent skills as a professional employee or a job candidate among a pool of human experts? This question becomes particularly intriguing when considering the nominal $20 per month charge for priority use of the OpenAI API. A few hundred dollars for a relentless personal assistant could be worthwhile.

This study introduces a novel benchmark [52], the Professional Certification Benchmark Exam Dataset. This dataset is derived from a representative sample of over 1100 professional certification practice exams, encompassing 5197 questions [53]. Additionally, the benchmark incorporates synthetically constructed examples from 49 skill tests identified as challenging in human surveys [54]. These synthetic datasets propose 10-20 question panels, each with a separate session for answering (resulting in 940 sessions). This comprehensive panel includes 1149 occupational skill evaluations, ranging from sensory-intensive roles like wine sommelier to technically rigorous roles in offensive cybersecurity. The novel dataset [52] comprises 1149 exams, presenting 6137 multiple-choice questions for the AI model to tackle.

This research extends upon previous internal assessments by OpenAI of large language models [4], primarily focusing on medical and legal professions (bar exams) and various scholastic achievement or advanced placement exams. A key objective of these assessments is to track the evolution of the general "zero-shot" knowledge [18] as the large language model (LLM) advances in terms of sophistication and capabilities. Here, "zero-shot" refers to the LLM's ability to answer questions without any prior example cases. This case that lacks any preceding examples is compared with "few-shot" prompt engineering, akin to a human candidate who has done some preparatory studying. In parallel with evaluating the capabilities of large language models (LLMs) in professional exam scenarios, there's a compelling need to understand how these models compare to human proficiency [4]. This study further explores the methodology and results of a comprehensive human comparison study to fulfill this objective.

## 2. METHODS

The study divided the research methodology into three stages: assembling a professional certification dataset at scale, testing multiple model variants, and comparing these models with human volunteers.

The first stage involves collecting and consolidating a large-scale professional certification dataset. This dataset encompasses an extensive range of practice tests, including those certifications typically evaluated through multiple-choice questions. The diversity of the dataset ensures a broad spectrum of professional roles, from wine sommeliers to certified accountants, are covered.

Various model variants are subjected to this robust testing regime in the second stage. This step allows us to assess the progress made over the last two years in developing AI models that can confidently tackle professional exams, thus opening the possibility of their deployment in many professional roles.

Finally, in the third stage, the performance of these AI models is compared with human volunteers. This comparative analysis provides a comprehensive understanding of how these AI models fare against human expertise, paving the way for future improvements and refinements in AI capabilities.

## 2.1 Benchmark Professional Certification Dataset

This study delves into an extensive collection of 1149 publicly available professional certification practice tests offering various specializations. The methodology for assembling this benchmark dataset involves a two-pronged approach.

Firstly, we comprehensively review all 1100 practice exams across various professions [53]. This review includes a cursory exploration of 49 certifications, providing an overview of the breadth of knowledge domains covered by these tests.

Secondly, we undertake a detailed study of a cybersecurity cloud certification. This certification typically mandates a six-month preparatory period for prospective applicants. By completing this certification, we aim to illustrate the depth of knowledge required in some of these specialized fields.

Thirdly, we examine the potential of using large language models (LLMs) to pass more than forty flight and aircraft maintenance tests administered by professional certification organizations and the US Federal Aviation Administration (FAA). We validate eight LLMs as closed (Open AI GPT series) and open-source (Meta Llama-2) models. We assess their advisory application at scale in aviation training and the implications of using such technology to ensure the safety and reliability of airborne operations. The 41 flight tests evaluate the competency of pilots, mechanics, and air traffic controllers, ensuring that they can safely operate aircraft in varying conditions, potentially with an AI co-pilot providing competent advice.

All the tests in the dataset are in the multiple-choice format, a preferred standard in many professional certifications due to its amenability to automated and objective grading methods. This diverse and comprehensive dataset is a robust benchmark for assessing the capabilities of various AI models in a wide range of professional domains.

## 2.2 Model Evaluation and Comparisons

The evaluation process in this study involves a comparative analysis of at least four OpenAI models released as API interfaces since 2021. These models, named alphabetically after renowned innovators - Ada (Lovelace), (Charles) Babbage, (Marie) Curie, and (Leonardo) DaVinci - represent different stages of AI development, with DaVinci being the most advanced model available in 2022, preceding the upgrade from GPT3 to GPT4. As of May 2023, OpenAI has not included GPT4 in its API formats, with the latest high-performing model being labeled as gpt3.5-turbo.

The metrics collected for each model include the model type, the score achieved on multiple-choice questions, and whether the model passed the professional certification exam with an accuracy of over 70%. The testing approach for all models was "zero-shot" learning, implying no specific examples of answering multiple-choice questions were provided to the models before the final prompt. Based on our observations, the "few-shot" approach, providing examples, proved unnecessary as all the tested models correctly followed instructions to choose a letter answer (A-D) representing the model's response.

For all models evaluated, the prompt included an instruction such as "You are an expert in professional certifications in <<BLANK>>. Answer the multiple-choice questions using the format Answer: [insert]". In this prompt, <<BLANK>> represented the broad category of the exam, such as "NIST SP800-171 CMMC Certification". For models where the API allows adjustment of the creativity level or temperature, we set this value to 0. This approach aimed to limit potential drift in the model's responses, aligning with other OpenAI applications focusing on question-answer interactions rather than creative outputs.

## 2.3 Human Trials and Crowdsourcing

The study was designed to compare human participants and various LLMs, including GPT3.5, GPT3, Anthropic, and OpenAssistant. The comparison study comprised a series of professional certification exams drawn from the same pool as those used to assess the LLMs. This approach ensured a fair and direct comparison, with humans and LLMs evaluated against the same criteria.

A total of 35 human participants were selected to represent a diverse range of professional backgrounds and expertise levels. This diverse group ensured a wide breadth of knowledge and skills, facilitating a more comprehensive comparison with the LLMs.

To enable an equitable comparison between human participants and LLMs, we randomly selected a subset of 100 questions from our extended Professional Certification Benchmark Exam Dataset [52], which comprises over 6,000 questions. This random selection ensured a fair representation of the dataset's diversity.

Our study involved two groups of human participants: a) Group 1 consisted of ten individuals from our research team's network, featuring a diverse mix of domain experts and individuals with general knowledge. b) Group 2: This group comprised twenty-five remunerated click workers [56], selected based on their interest and willingness to participate in the study. The crowdsourcing was limited to English-speaking participants self-selected from the more than four million site members. Upon completing the 100-question sub-sample, each crowdsourced participant received two US dollars, while Group 1 received no compensation.

Each participant was given a set of 100 randomly selected multiple choice questions to answer, replicating an actual exam scenario. The responses from the human participants were then compared to the answers generated by the LLMs for the same set of questions. This method directly compared performance between human participants and AI models.

## 3. RESULTS

Table 1 summarizes the results of five representative professional certifications administered to two AI models, GPT3 and Turbo-GPT3.5. A comprehensive scoresheet for all 1149 professional certifications is available in the Appendix. Figure 1 illustrates the performance difference between the two models and ranks them based on the number of certifications in which they achieved a passing grade (>70% correct) using the zero-shot approach.

GPT3 successfully passed approximately 39% of the diverse professional certifications without specific fine-tuning or exam preparation. The certifications primarily encompassed various computer-related vocations, spanning from high-value cybersecurity certifications to project management. Unlike previous studies that chose exams based on academic performance, this benchmark dataset emphasizes vocational readiness. Notably, over 100 cybersecurity certifications typically recommend a preparation time of at least six months for human applicants. One significant certification, the Offensive Security Certified Professional (OSCP), is a valuable skill for penetration testers that often leads to substantial salary boosts in industry and government roles. Remarkably, GPT3.5 scored a perfect 100% on the OSCP exam.

The benchmark test [52] encompasses a broad range of academic qualifications necessary for professional employability, from the high school graduation exam (General Education Development (GED)) to graduate admissions (Graduate Management Admission Test (GMAT)). In another notable category, the model demonstrated competent skills in nursing (Test of Essential Academic Skills, or TEAS), licensed counseling (National Counselor Examination (NCE)), and pharmacy (North American Pharmacist Licensure Examination, NAPLEX), consistent with previous testing [4]. The model also passed the teacher certification exam (PRAXIS), a crucial requirement for employment in US public education. Beyond the medical or legal bar exams, the Financial Industry Regulatory Authority (FINRA) Series 6 exam, a lucrative opportunity for financial advisors with a human pass rate of 58% [55], was passed by GPT-3.5 with a score of 70% without any preparation.

As evidenced by the tabulated professional certification results (see Appendix), the latest LLMs show qualifications in various computer-related fields. These include cloud and virtualization (Amazon AWS, Alibaba, Google, Microsoft, VMWare), business analytics (Project Management Institute, Six Sigma, Configuration Management), cybersecurity (CompTIA, EC Council, InfoSec Institute, IAPP, Solarwinds, Palo Alto Networks, Blockchain, etc.), network setup and repair (Dell, HP, Citrix, Novell, Fortinet), and data analytics (Tableau, Qlikview Developer, UIPath).

| Table 1. Example Scores Based on Five Professional Examinations Administered to Two Models | | |
|---|---|---|
| Certification | Turbo-GPT3.5 | GPT3 |
| ALIBABA ACA CLOUD1 | 100% | 100% |
| COMPTIA 220 1102 | 100% | 100% |
| COMPTIA 220 902 | 100% | 100% |
| CompTIA XK0 005 | 100% | 100% |
| CSA CCSK | 100% | 100% |

Ironically, a chat interface (turbo-gpt3.5) scores well on traditional customer service responsibilities tested by Avaya exams, a likely early destination for human augmentation with chatbots for routine advice and call centers. Interestingly, as evaluated by Avaya exams, Turbo-GPT3.5, a chat interface model, exhibits strong performance in traditional customer service responsibilities. This success suggests a promising avenue for human augmentation via chatbots in routine advisory roles and call centers.

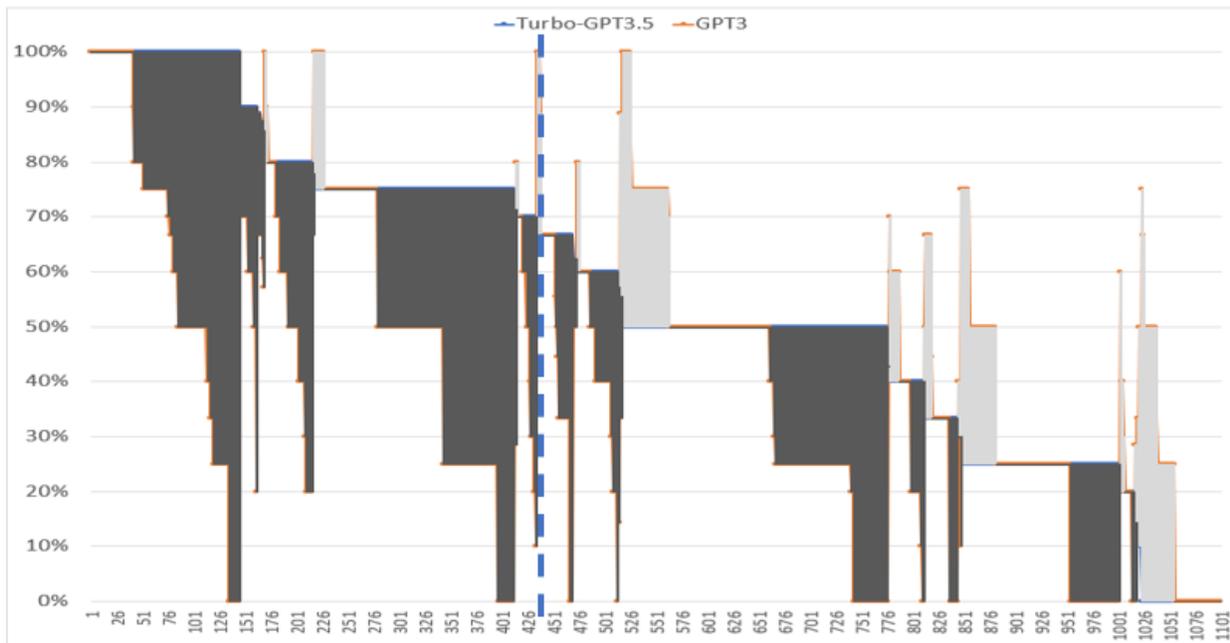

*Figure 1. Rank ordered professional exam scores between two LLM models, rated by the percentage of questions answered correctly. The blue dotted line shows the 435 examinations scoring 70% or higher with zero-shot approach.*

However, not all tests resulted in success. Notable certification failures occurred in areas where successes were also recorded, including VMWare (virtualization), teaching writing (Praxis), and logical legal reasoning (LSAT Section 1, Logical Reasoning). These results indicate the need for further investigation to ascertain whether the format of these multiple-choice exams is conducive to prompting LLMs and whether the test evaluation methods are suitable for automated scoring. For instance, low scores in writing and Python coding could potentially stem from prompt ambiguity in the question submission, given that LLMs have previously demonstrated high competencies in grammar, language comprehension, and coding [4].

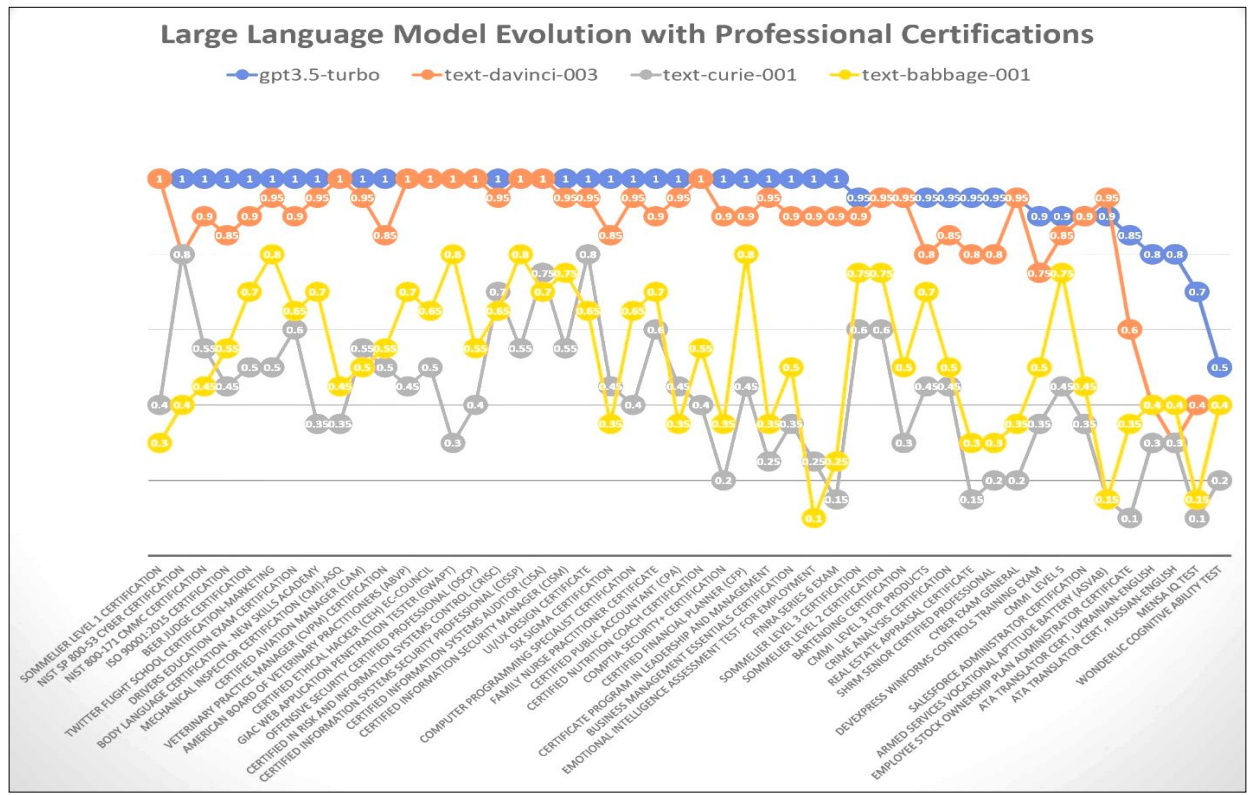

*Figure 2. Comparison of hardest certifications across different model evolutions, including experience-based tests*

Figure 2 presents a challenging subset of non-computer-related tasks, such as wine selection (sommelier exams), language translation, driver's education, and various IQ tests (Mensa, Wonderlic). The motivation for selecting these particular tests was their high ranking as the most challenging certifications [54]. They encompass a variety of professions, including accountants (CPA), veterinarians (CVPM), aviation inspectors (CMI-ASQ, CAM), real estate appraisers, human resources professionals (SHRM), and financial planners (CFP).

Another consideration was the inclusion of exams that rely on sensory or emotional experience, aspects that LLMs can only understand through acquired knowledge, not direct experience. These tests include wine and beer tasting, emotional intelligence assessment (EQ), body language interpretation, and driver's education.

Figure 2 demonstrates that Turbo-GPT3.5 achieved passing grades (>70%) in all these tests except for Wonderlic, which scored 50%. For reference, the average human score on the Wonderlic test is 40% (20/50 correct). Interestingly, the results show no significant bias in experience-based tests such as sensory or emotional evaluations. This result can likely be attributed to the many internet pages used to train and fine-tune the models, which include numerous text-generation examples of these scenarios.

The progression from OpenAI's Babbage to Turbo model showed a median improvement of 60% in graded performance (100% in Turbo vs. 40% in Babbage). Achieved in less than a few years, this advancement implies that, with targeted attention to the latest model's shortcomings as highlighted in the Appendix examples, a highly efficient model capable of acing the most demanding professional certifications could be developed.

One of the fascinating revelations from the human comparison study was the significant variability in human performance. Scores varied widely, ranging from a high of 92 (similar to top-performing LLMs) to a low of 23 (comparable to lower-performing LLMs). The first cohort of 12 colleagues scored an average of 73.7, which ranks below GPT-3. The second cohort of 25 paid participants on clickworkers scored an average of 30 points lower, or 43.4, which in multiple choice questions is closer to random (25) than the colleague pool. This vast range of performance highlights human evaluations' subjective nature, which could introduce biases and inconsistencies in human responses and self-selection for testing generally..

The demonstrate LLM use cases to tutor prospective aviation professionals, providing information, answering queries, and simulating scenarios to test their theoretical knowledge (Figure 3). Furthermore, they can generate study guides tailored to individual learning patterns. The study directly addresses the significant concerns about hallucinations and the reliability of the information provided by the LLM. We compare the passing grades of GPT-4 (100% passing) to Open AI's Curie 2020 model (0% passing). One surprising result is that 2023 open-source models like Llama-2 exceed the closed-source state-of-the-art models from the previous three years.

In a comprehensive analysis (Figure 3), we juxtaposed the performance of human participants with that of the LLMs, specifically GPT3.5, GPT3, Anthropic, and OpenAssistant. The primary aim of this comparison was to assess the degree to which LLMs either surpassed or lagged behind their human counterparts. This evaluation provided a measure of LLM performance relative to human expertise and valuable insights into areas where LLMs excelled or required further enhancement.

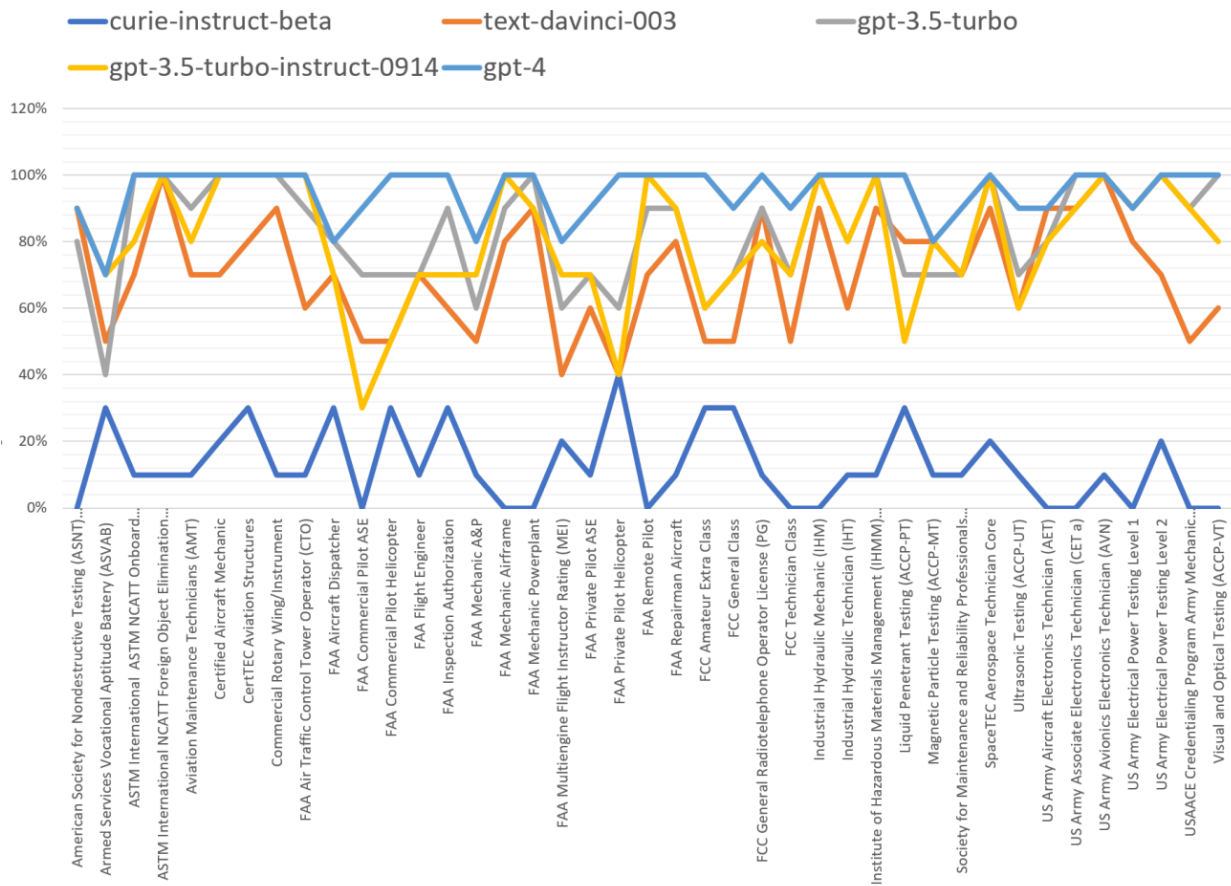

*Figure 3. GPT Series results for 41 Aviation Related Certifications and Exams*

## 4. DISCUSSION

This study focuses on three critical aspects of AI: reliability, competencies, and trustworthiness outside of their specific training sets. This triad highlights the inherent trade-offs and concerns surrounding the development of large language models (LLMs) that lack a verifiable world model grounded in physics, common sense, or empathy. Recent efforts to manipulate model behavior through prompt engineering further accentuate these concerns, as the model's response is often entirely contingent on the prompt, irrespective of whether the user's intent is nefarious, malicious, or benign. These safety issues fall into three categories: constraints on reasoning ability, hallucination, and user interactivity.

In this study, we assembled a comprehensive professional certification benchmark and evaluated the competency and capabilities of various LLMs without any preparation or specific attention to training examples. The most recent models, such as Turbo-GPT3.5, demonstrated competencies spanning multiple medical, legal, technical, and financial domains, thus reflecting a broad spectrum of human expertise. The study's original contribution lies in creating this benchmark dataset, allowing for future testing and comparing new models across various capabilities. The OpenAI API offers a scalable approach to expanding this dataset from its 5197 questions to a broader array of professional competencies.

These results provide valuable insights into AI trust and confidence limits, particularly during this critical phase of heightened focus on AI safety. Our work complements similar quantitative evaluations provided by OpenAI (known as eval corpus [51]) and recent studies illustrating how models like ChatGPT and GPT-4 surpass their predecessors in exam-taking tasks across a variety of educational milestones [19].

OpenAI has presented data comparing a single GPT generation (3/4) and its progress in taking Advanced Placement (high school) exams, legal and medical bar exams, the Graduate Record Examination (GRE), the Scholastic Aptitude Test (SAT), and various competitive math and biology Olympiads [19]. However, there has been criticism [40] of the testing industry's shift towards multiple-choice formats due to their ease of automated scoring. Such a feature may inadvertently assist LLMs in guessing well when considering the next token prediction [2-3].

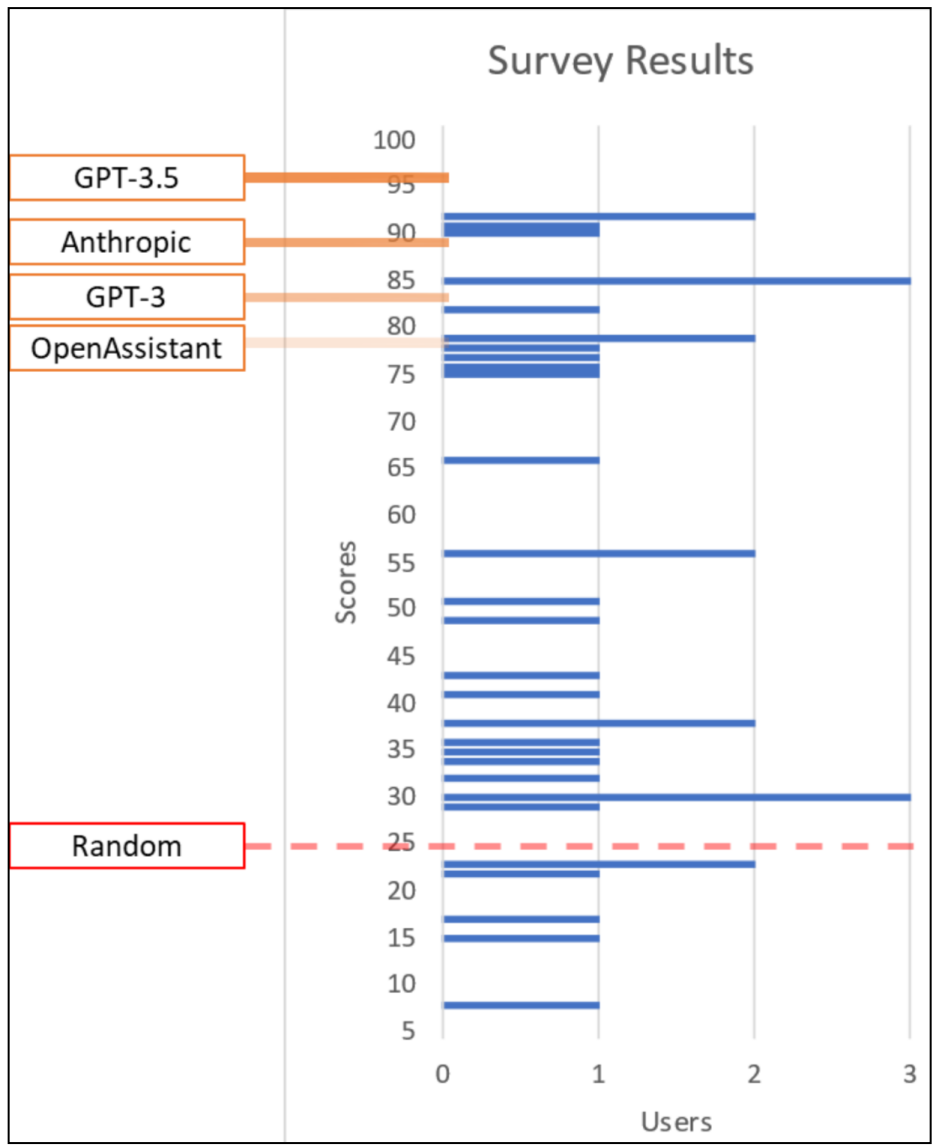

*Figure 4. Comparative distribution of scores between human and AI models*

While multiple-choice tests provide a practical method for assessing professional skills, they also have limitations. For instance, they may not capture the depth of understanding or the ability to apply knowledge in real-world situations, which are critical aspects of professional competence. Furthermore, they may inadvertently favor LLMs, which are adept at processing large amounts of information and identifying patterns, advantages that may not reflect a genuine understanding of the subject matter.

Compared with human trials, our results reveal a considerable range of variability in human performance, highlighting the subjective nature of human evaluations (Figure 4). The substantial variability observed in human performance underscores key considerations for integrating LLMs into professional certification processes. Human evaluation and scoring can be influenced by a wide array of factors, including individual knowledge, personal experience, and subjective judgment, leading to potential variations in scoring. Given this, there is a pressing need to implement rigorous standardization measures and robust evaluation protocols when leveraging AI models for professional certification.

The comparative analysis between human participants and LLMs offers valuable insights into the potential role of AI in the professional certification process. While LLMs demonstrate remarkable capabilities in successfully passing professional exams, understanding the domains where human expertise still surpasses AI performance is critical. This knowledge can help refine and enhance the use of AI in certification, ensuring that it complements human expertise rather than seeking to replace it. This approach would create a more collaborative certification process, leveraging AI's and human evaluators' strengths for a more comprehensive and practical assessment of professional competencies.

Despite these limitations, our findings provide valuable insights into the potential of LLMs to perform tasks that require a broad range of professional knowledge and skills. They highlight areas where LLMs excel and need further refinement, providing a roadmap for future AI development.

## 5. CONCLUSIONS AND FUTURE WORK

This study's original contributions lie in creating a comprehensive benchmark for professional certifications, providing a comparative analysis of LLM and human performance, and offering valuable insights into the potential and limitations of AI in the context of professional competency evaluations. These findings provide a strong foundation for future research in this rapidly evolving field.

One of the intriguing findings of this study lies in the LLMs' performance in more experience-based testing, which helps reduce the influence of data parroting from the model's training phase. A particularly noteworthy aspect of this research was testing sensory and emotional qualifications. Contrary to widely held beliefs that the current generation of AI would struggle to interact naturally with humans, the results showed that these models could engage in emotionally charged and seemingly sensory-dependent evaluations in a quantifiable conversational context.

While traditional educational evaluations assess verbal, comprehension, and mathematical understanding as core skills of general education, professional certifications go a step further. They offer assurance that a candidate, whether a human or an AI model, possesses the requisite knowledge and skills to perform competently in specific fields.

Areas of future interest may include further investigations into experience-based testing contributions, refining methods for evaluating models' reasoning abilities and exploring novel ways of testing the models' capabilities beyond simply recalling information. A safeguarding test involves question rephrasing and negation examples that might reduce the risk that specific online questions appear in training data for such comprehensive LLMs.

A critical concern when evaluating LLM performance like GPT3 and Turbo-GPT3.5 on professional certification exams is the risk of data leakage. Data leakage refers to the possibility that the model may have encountered the exact or similar questions during its training phase. Given that these LLMs are trained on a vast corpus of text data from the internet, there is a chance that they might have been exposed to similar questions or related material, thereby influencing their performance on the exams.

LLMs are adept at detecting patterns in data and can leverage any information available during training. If a question or a similar one appeared in the training data, the model might remember and recall the associated answer rather than reasoning it out based on its understanding. This 'recall' capability can lead to inflated performance results, which may not accurately represent the model's competency in a given domain.

Similarly, small changes in question-wording can also significantly impact the model's performance. These models are sensitive to input phrasing and can interpret slight modifications in a question as an entirely different query. This sensitivity can lead to variability in performance based on how questions are framed, which could affect the model's ability to pass a given exam consistently.

These concerns emphasize the need to design and interpret professional certification exams for LLMs carefully. It also underscores the importance of developing more sophisticated methods for evaluating these models, such as using novel questions not present in the training data or creating exams that test the model's ability to reason and understand rather than recall information. Moreover, it reinforces the necessity of further research into understanding the behavior of these models and developing mechanisms to ensure their reliable and fair evaluation.

## ACKNOWLEDGMENTS

The author would like to thank the PeopleTec Technical Fellows program for its encouragement and project assistance. The author thanks the researchers at Open AI for developing large language models and allowing public access to ChatGPT.The author would like to thank the PeopleTec Technical Fellows program for its encouragement and project assistance. The author thanks the researchers at Open AI for developing large language models and allowing public access to ChatGPT.

## REFERENCES

[1] Johnson, K. E., & Ma, P. (1999). Understanding language teaching: Reasoning in action. Boston, MA: Heinle & Heinle.

[2] Kejriwal, M., Santos, H., Mulvehill, A. M., & McGuinness, D. L. (2022). Designing a strong test for measuring true common-sense reasoning. Nature Machine Intelligence, 4(4), 318-322.

[3] Radford, A., Narasimhan, K., Salimans, T., & Sutskever, I. (2018). Improving language understanding by generative pre-training. OpenAI, https://cdn.openai.com/research-covers/language-unsupervised/language_understanding_paper.pdf

[4] OpenAI, (2023) GPT-4 Technical Report, https://arxiv.org/abs/2303.08774

[5] Bubeck, S., Chandrasekaran, V., Eldan, R., Gehrke, J., Horvitz, E., Kamar, E., ... & Zhang, Y. (2023). Sparks of artificial general intelligence: Early experiments with gpt-4. arXiv preprint arXiv:2303.12712.

[6] Kung, T. H., Cheatham, M., Medenilla, A., Sillos, C., De Leon, L., Elepaño, C., ... & Tseng, V. (2023). Performance of ChatGPT on USMLE: Potential for AI-assisted medical education using large language models. PLoS digital health, 2(2), e0000198.

[7] Gilson, A., Safranek, C. W., Huang, T., Socrates, V., Chi, L., Taylor, R. A., & Chartash, D. (2023). How does CHATGPT perform on the United States Medical Licensing Examination? the implications of large language models for medical education and knowledge assessment. JMIR Medical Education, 9(1), e45312.

[8]     Jiao, W., Wang, W., Huang, J. T., Wang, X., & Tu, Z. (2023). Is ChatGPT a good translator? A preliminary study. arXiv preprint arXiv:2301.08745.

[9]     Bang, Y., Cahyawijaya, S., Lee, N., Dai, W., Su, D., Wilie, B., ... & Fung, P. (2023). A multitask, multilingual, multimodal evaluation of chatgpt on reasoning, hallucination, and interactivity. arXiv preprint arXiv:2302.04023.

[10]    Noever, D., Kalin, J., Ciolino, M., Hambrick, D., & Dozier, G. (2021). Local translation services for neglected languages. *arXiv preprint arXiv:2101.01628*.

[11]    Qin, C., Zhang, A., Zhang, Z., Chen, J., Yasunaga, M., & Yang, D. (2023). Is chatgpt a general-purpose natural language processing task solver?. arXiv preprint arXiv:2302.06476.

[12]    Zhang, C., Zhang, C., Zheng, S., Qiao, Y., Li, C., Zhang, M., ... & Hong, C. S. (2023). A Complete Survey on Generative AI (AIGC): Is ChatGPT from GPT-4 to GPT-5 All You Need?. arXiv preprint arXiv:2303.11717.

[13]    Bommarito II, M., & Katz, D. M. (2022). GPT Takes the Bar Exam. arXiv preprint arXiv:2212.14402.

[14]    Surameery, N. M. S., & Shakor, M. Y. (2023). Use chat gpt to solve programming bugs. International Journal of Information Technology & Computer Engineering (IJITC) ISSN: 2455-5290, 3(01), 17-22.

[15]    Fijačko, N., Gosak, L., Štiglic, G., Picard, C. T., & Douma, M. J. (2023). Can ChatGPT pass the life support exams without entering the American heart association course?. Resuscitation, 185.

[16]    Frieder, S., Pinchetti, L., Griffiths, R. R., Salvatori, T., Lukasiewicz, T., Petersen, P. C., ... & Berner, J. (2023). Mathematical capabilities of chatgpt. arXiv preprint arXiv:2301.13867.

[17]    Mbakwe, A. B., Lourentzou, I., Celi, L. A., Mechanic, O. J., & Dagan, A. (2023). ChatGPT passing USMLE shines a spotlight on the flaws of medical education. PLOS Digital Health, 2(2), e0000205.

[18]    Srivastava, P., Ganu, T., & Guha, S. (2022). Towards Zero-Shot and Few-Shot Table Question Answering using GPT-3. arXiv preprint arXiv:2210.17284.

[19]    Kocoń, J., Cichecki, I., Kaszyca, O., Kochanek, M., Szydło, D., Baran, J., ... & Kazienko, P. (2023). ChatGPT: Jack of all trades, master of none. arXiv preprint arXiv:2302.10724.

[20]    Kung, T. H., Cheatham, M., Medenilla, A., Sillos, C., De Leon, L., & Elepaño, C. (2023). Performance of ChatGPT on USMLE: Potential for AI-assisted medical education using large language models. PLOS Digit Health 2 (2): e0000198.

[21]    Peng, K., Ding, L., Zhong, Q., Shen, L., Liu, X., Zhang, M., ... & Tao, D. (2023). Towards Making the Most of ChatGPT for Machine Translation. arXiv preprint arXiv:2303.13780.

[22]    Felten, E., Raj, M., & Seamans, R. (2023). How will Language Modelers like ChatGPT Affect Occupations and Industries?. arXiv preprint arXiv:2303.01157.

[23]    Gilson, A., Safranek, C., Huang, T., Socrates, V., Chi, L., Taylor, R. A., & Chartash, D. (2022). How Well Does ChatGPT Do When Taking the Medical Licensing Exams? The Implications of Large Language Models for Medical Education and Knowledge Assessment. medRxiv, 2022-12.

[24]    Das, D., Kumar, N., Longjam, L. A., Sinha, R., Roy, A. D., Mondal, H., & Gupta, P. (2023). Assessing the Capability of ChatGPT in Answering First-and Second-Order Knowledge Questions on Microbiology as per Competency-Based Medical Education Curriculum. Cureus, 15(3).

[25]    Nori, H., King, N., McKinney, S. M., Carignan, D., & Horvitz, E. (2023). Capabilities of gpt-4 on medical challenge problems. *arXiv preprint arXiv:2303.13375*.

[26]    Liu, H., Ning, R., Teng, Z., Liu, J., Zhou, Q., & Zhang, Y. (2023). Evaluating the Logical Reasoning Ability of ChatGPT and GPT-4. arXiv preprint arXiv:2304.03439.

[27]    Fleming, S. L., Morse, K., Kumar, A. M., Chiang, C. C., Patel, B., Brunskill, E. P., & Shah, N. (2023). Assessing the Potential of USMLE-Like Exam Questions Generated by GPT-4. medRxiv, 2023-04.

[28] Nunes, D., Primi, R., Pires, R., Lotufo, R., & Nogueira, R. (2023). Evaluating GPT-3.5 and GPT-4 Models on Brazilian University Admission Exams. arXiv preprint arXiv:2303.17003.

[29] Jiao, W. X., Wang, W. X., Huang, J. T., Wang, X., & Tu, Z. P. (2023). Is ChatGPT a good translator? Yes with GPT-4 as the engine. arXiv preprint.

[30] Ali, R., Tang, O. Y., Connolly, I. D., Zadnik Sullivan, P. L., Shin, J. H., Fridley, J. S., ... & Telfeian, A. E. (2023). Performance of ChatGPT and GPT-4 on Neurosurgery Written Board Examinations. medRxiv, 2023-03.

[31] Holmes, J., Liu, Z., Zhang, L., Ding, Y., Sio, T. T., McGee, L. A., ... & Liu, W. (2023). Evaluating large language models on a highly-specialized topic, radiation oncology physics. arXiv preprint arXiv:2304.01938.

[32] Singh, M., SB, V., & Malviya, N. (2023). Mind meets machine: Unravelling GPT-4's cognitive psychology. arXiv preprint arXiv:2303.11436.

[33] Chang, K. K., Cramer, M., Soni, S., & Bamman, D. (2023). Speak, Memory: An Archaeology of Books Known to ChatGPT/GPT-4. arXiv preprint arXiv:2305.00118.

[34] Pursnani, V., Sermet, Y., & Demir, I. (2023). Performance of ChatGPT on the US Fundamentals of Engineering Exam: Comprehensive Assessment of Proficiency and Potential Implications for Professional Environmental Engineering Practice. arXiv preprint arXiv:2304.12198.

[35] Li, Y., & Duan, Y. (2023). The Evaluation of Experiments of Artificial General Intelligence with GPT-4 Based on DIKWP. arXiv preprint.

[36] Tanaka, Y., Nakata, T., Aiga, K., Etani, T., Muramatsu, R., Katagiri, S., ... & Nomura, A. (2023). Performance of Generative Pretrained Transformer on the National Medical Licensing Examination in Japan. medRxiv, 2023-04.

[37] Tang, F., & Duan, Y. The Capability Evaluation of GPT-4 and Tongyi Qianwen on Financial Domain with DIKWP Analysis.

[38] Zhong, W., Cui, R., Guo, Y., Liang, Y., Lu, S., Wang, Y., ... & Duan, N. (2023). AGIEval: A Human-Centric Benchmark for Evaluating Foundation Models. arXiv preprint arXiv:2304.06364.

[39] King, M. (2023). Administration of the text-based portions of a general IQ test to five different large language models.

[40] Larsen, S. K. Creating Large Language Model Resistant Exams: Guidelines and Strategies. https://arxiv.org/ftp/arxiv/papers/2304/2304.12203.pdf

[41] Yuan, Z., Yuan, H., Tan, C., Wang, W., & Huang, S. (2023). How well do Large Language Models perform in Arithmetic tasks?. arXiv preprint arXiv:2304.02015.

[42] Rudolph, J., Tan, S., & Tan, S. War of the chatbots: Bard, Bing Chat, ChatGPT, Ernie and beyond. The new AI gold rush and its impact on higher education. Journal of Applied Learning and Teaching, 6(1).

[43] Küchemann, S., Steinert, S., Revenga, N., Schweinberger, M., Dinc, Y., Avila, K. E., & Kuhn, J. (2023). Physics task development of prospective physics teachers using ChatGPT. arXiv preprint arXiv:2304.10014.

[44] Jiao, W., Huang, J. T., Wang, W., Wang, X., Shi, S., & Tu, Z. (2023). Parrot: Translating during chat using large language models. arXiv preprint arXiv:2304.02426.

[45] Zhang, C., Zhang, C., Zheng, S., Qiao, Y., Li, C., Zhang, M., ... & Hong, C. S. (2023). A Complete Survey on Generative AI (AIGC): Is ChatGPT from GPT-4 to GPT-5 All You Need?. arXiv preprint arXiv:2303.11717.

[46] Borji, A. (2023). A categorical archive of ChatGPT failures. arXiv preprint arXiv:2302.03494.

[47] Li, M., Song, F., Yu, B., Yu, H., Li, Z., Huang, F., & Li, Y. (2023). API-Bank: A Benchmark for Tool-Augmented LLMs. arXiv preprint arXiv:2304.08244.

[48] Noever, D., & Ciolino, M. (2022). The Turing Deception. arXiv preprint arXiv:2212.06721.

[49] French, R. M. (2000). The Turing Test: the first 50 years. Trends in cognitive sciences, 4(3), 115-122.

[50] Sun, H., Lin, Z., Zheng, C., Liu, S., & Huang, M. (2021). Psyqa: A chinese dataset for generating long counseling text for mental health support. arXiv preprint arXiv:2106.01702.

[51] OpenAI (2023), Evals, https://github.com/openai/evals

[52] Noever, D. (2023), Professional Certification Benchmark Exam Dataset, https://github.com/reveondivad/certify

[53] Exam Topics, (2023), Actual Exam Material: A Community You Can Belong To, https://www.examtopics.com/

[54] Erudera, (2023), Top Toughest Exams in the World 2023, https://erudera.com/resources/top-toughest-exams-in-the-world/

[55] CPA Accounting Institute for Success (2022), Breakdown: Series 6 Pass/Fail Rate, https://www.ais-cpa.com/breakdown-series-6-pass-fail-rate

[56] Clickworker.com (2023), https://marketplace.clickworker.com/en/marketplace/home

## Authors

**David Noever** has research experience with NASA and the Department of Defense in machine learning and data mining. He received his BS from Princeton University and his Ph.D. from Oxford University as a Rhodes Scholar in theoretical physics.

**Matt Ciolino** has research experience in deep learning and computer vision. He received his Bachelor's in Mechanical Engineering from Lehigh University. Matt is pursuing graduate study in computer vision and machine learning at Georgia Tech.

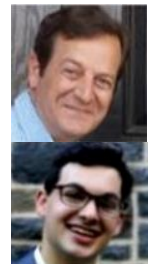

# Appendix A: Examination Evaluation Compared between Turbo GPT 3.5 vs. GPT 3

| Certification | Turbo-GPT3.5 | GPT3 | Certification | Turbo-GPT3.5 | GPT3 |
|---|---|---|---|---|---|
| ALIBABA ACA CLOUD1 | 100% | 100% | AAFM INDIA CWM LEVEL 2 | 50% | 50% |
| COMPTIA 220 1102 | 100% | 100% | AMAZON AWS CERTIFIED DEVELOPER ASSOCIATE DVA C02 | 50% | 50% |
| COMPTIA 220 902 | 100% | 100% | AMAZON AWS CERTIFIED DEVOPS ENGINEER PROFESSIONAL DOP C02 | 50% | 50% |
| COMPTIA XK0 005 | 100% | 100% | AMAZON AWS CERTIFIED SOLUTIONS ARCHITECT ASSOCIATE SAA C03 | 50% | 50% |
| CSA CCSK | 100% | 100% | AMAZON AWS CERTIFIED SOLUTIONS ARCHITECT PROFESSIONAL | 50% | 50% |
| DELL DEA 2TT4 | 100% | 100% | ANDROIDATC AND 402 | 50% | 50% |
| DMI PDDM | 100% | 100% | ARUBA ACCP V62 | 50% | 50% |
| ECCOUNCIL 212 82 | 100% | 100% | AVAYA 3300 | 50% | 50% |
| EXIN MORF | 100% | 100% | AVAYA 3308 | 50% | 50% |
| GENESYS GE0 803 | 100% | 100% | AVAYA 6202 | 50% | 50% |
| GOOGLE CLOUD DIGITAL LEADER | 100% | 100% | AVAYA 7391X | 50% | 50% |
| GOOGLE GOOGLE ANALYTICS | 100% | 100% | BCS FCBA | 50% | 50% |
| GOOGLE PROFESSIONAL DATA ENGINEER | 100% | 100% | BCS ISEB SWT2 | 50% | 50% |
| IAPP CIPT | 100% | 100% | BCS RE18 | 50% | 50% |
| LPI 102 500 | 100% | 100% | CERTNEXUS ITS 110 | 50% | 50% |
| MICROSOFT 62 193 | 100% | 100% | CHECKPOINT 156 560 | 50% | 50% |
| MICROSOFT 70 334 | 100% | 100% | CITRIX 1Y0 403 | 50% | 50% |
| MICROSOFT 70 398 | 100% | 100% | COMPTIA PT1 002 | 50% | 50% |
| MICROSOFT 70 465 | 100% | 100% | CWNP CWDP 303 | 50% | 50% |
| MICROSOFT 70 475 | 100% | 100% | CYBERARK CAU201 | 50% | 50% |
| MICROSOFT 70 488 | 100% | 100% | DATABRICKS CERTIFIED DATA ENGINEER ASSOCIATE | 50% | 50% |
| MICROSOFT 70 489 | 100% | 100% | DELL DEA 1TT4 | 50% | 50% |
| MICROSOFT 98 349 | 100% | 100% | DELL DEA 1TT5 | 50% | 50% |
| MICROSOFT 98 361 | 100% | 100% | DELL DEA 41T1 | 50% | 50% |
| MICROSOFT 98 366 | 100% | 100% | DELL DEA 64T1 | 50% | 50% |
| MICROSOFT 98 382 | 100% | 100% | DELL DES 2T13 | 50% | 50% |
| MICROSOFT AI 102 | 100% | 100% | DELL DES 5221 | 50% | 50% |
| MICROSOFT AI 900 | 100% | 100% | DELL DES DD23 | 50% | 50% |
| MICROSOFT AZ 101 | 100% | 100% | DELL E20 368 | 50% | 50% |

| Certification | Turbo-GPT3.5 | GPT3 | Certification | Turbo-GPT3.5 | GPT3 |
|---|---|---|---|---|---|
| MICROSOFT AZ 140 | 100% | 100% | ECCOUNCIL 312 39 | 50% | 50% |
| MICROSOFT AZ 200 | 100% | 100% | ECCOUNCIL 312 50V12 | 50% | 50% |
| MICROSOFT AZ 801 | 100% | 100% | EXIN ASM | 50% | 50% |
| MICROSOFT MB 900 | 100% | 100% | FORTINET NSE4 FGT 60 | 50% | 50% |
| MICROSOFT MB 901 | 100% | 100% | FORTINET NSE5 EDR 5 0 | 50% | 50% |
| MICROSOFT MS 700 | 100% | 100% | FORTINET NSE5 FCT 7 0 | 50% | 50% |
| PMI PMI PBA | 100% | 100% | FORTINET NSE7 SDW 64 | 50% | 50% |
| SALESFORCE DEV 501 | 100% | 100% | GAQM CBAF 001 | 50% | 50% |
| SNIA S10 110 | 100% | 100% | GIAC GSNA | 50% | 50% |
| TEST PREP MCAT SECTION 1 VERBAL REASONING | 100% | 100% | GOOGLE GSUITE | 50% | 50% |
| TEST PREP TEAS SECTION 4 SENTENCE COMPLETION | 100% | 100% | GUIDANCE SOFTWARE GD0 110 | 50% | 50% |
| THE OPEN GROUP OG0 093 | 100% | 100% | HP HPE0 S51 | 50% | 50% |
| MICROSOFT 98 365 | 100% | 90% | HP HPE2 E67 | 50% | 50% |
| ANDROIDATC AND 401 | 100% | 80% | HP HPE2 T36 | 50% | 50% |
| COMPTIA PT0 002 | 100% | 80% | HUAWEI H12 811 | 50% | 50% |
| ECCOUNCIL 312 50V11 | 100% | 80% | IAPP CIPP A | 50% | 50% |
| GAQM CLSSGB | 100% | 80% | ISACA COBIT 2019 | 50% | 50% |
| IAPP CIPP E | 100% | 80% | ISQI CTAL TM SYLL2012 | 50% | 50% |
| ISC CISSP ISSAP | 100% | 80% | MAGENTO MAGENTO 2 CERTIFIED ASSOCIATE DEVELOPER | 50% | 50% |
| MICROSOFT 98 364 | 100% | 80% | MICROSOFT 70 332 | 50% | 50% |
| NACVA CVA | 100% | 80% | MICROSOFT 70 339 | 50% | 50% |
| NOKIA NOKIA 4A0 100 | 100% | 80% | MICROSOFT 70 342 | 50% | 50% |
| APICS CPIM BSP | 100% | 75% | MICROSOFT 70 348 | 50% | 50% |
| APPLE 9L0 012 | 100% | 75% | MICROSOFT 70 414 | 50% | 50% |
| ASQ CSSGB | 100% | 75% | MICROSOFT 70 487 | 50% | 50% |
| AVAYA 72200X | 100% | 75% | MICROSOFT 98 375 | 50% | 50% |
| EXIN ASF | 100% | 75% | MICROSOFT AI 100 | 50% | 50% |
| EXIN EX0 008 | 100% | 75% | MICROSOFT AZ 600 | 50% | 50% |
| EXIN EX0 115 | 100% | 75% | MICROSOFT DP 500 | 50% | 50% |
| FORTINET NSE5 FAZ 54 | 100% | 75% | MICROSOFT MB 330 | 50% | 50% |
| GAQM BPM 001 | 100% | 75% | MICROSOFT MB2 710 | 50% | 50% |
| GAQM CDCP 001 | 100% | 75% | MICROSOFT MB2 713 | 50% | 50% |
| GAQM CSM 001 | 100% | 75% | MICROSOFT MS 203 | 50% | 50% |
| GOOGLE PROFESSIONAL GOOGLE WORKSPACE ADMINISTRATOR | 100% | 75% | MICROSOFT MS 302 | 50% | 50% |

| Certification | Turbo-GPT3.5 | GPT3 | Certification | Turbo-GPT3.5 | GPT3 |
|---|---|---|---|---|---|
| IAAP CPACC | 100% | 75% | MICROSOFT PL 600 | 50% | 50% |
| IIBA IIBA AAC | 100% | 75% | NETAPP NS0 003 | 50% | 50% |
| ISACA COBIT 5 | 100% | 75% | NETAPP NS0 182 | 50% | 50% |
| ISQI CTFL | 100% | 75% | NETSUITE SUITEFOUNDATION CERTIFICATION EXAM | 50% | 50% |
| LPI 010 160 | 100% | 75% | NI CLAD | 50% | 50% |
| MICROSOFT 70 347 | 100% | 75% | NUTANIX NCP | 50% | 50% |
| MICROSOFT AZ 305 | 100% | 75% | PALO ALTO NETWORKS ACE | 50% | 50% |
| PALO ALTO NETWORKS PCCSA | 100% | 75% | PALO ALTO NETWORKS PCDRA | 50% | 50% |
| SIX SIGMA LSSYB | 100% | 75% | PEGASYSTEMS PEGAPCDC80V1 | 50% | 50% |
| SOA S9003 | 100% | 75% | RIVERBED 830 01 | 50% | 50% |
| TEST PREP GED SECTION 4 LANGUAGE ARTS READING | 100% | 75% | SALESFORCE CERTIFIED DATA ARCHITECT | 50% | 50% |
| WORLDATWORK C8 | 100% | 75% | SALESFORCE CERTIFIED OMNISTUDIO DEVELOPER | 50% | 50% |
| HRCI SPHR | 100% | 70% | SALESFORCE FIELD SERVICE LIGHTNING CONSULTANT | 50% | 50% |
| MICROSOFT 70 680 | 100% | 70% | SAP C S4CFI 2202 | 50% | 50% |
| ACFE CFE | 100% | 67% | SCRUM PSM II | 50% | 50% |
| MICROSOFT AZ 304 | 100% | 67% | SIX SIGMA ICYB | 50% | 50% |
| MICROSOFT SC 100 | 100% | 67% | SPLUNK SPLK 2002 | 50% | 50% |
| ALCATEL LUCENT 4A0 104 | 100% | 60% | SPLUNK SPLK 3001 | 50% | 50% |
| COMPTIA CAS 002 | 100% | 60% | SYMANTEC 250 428 | 50% | 50% |
| COMPTIA FC0 U51 | 100% | 60% | THE OPEN GROUP OG0 061 | 50% | 50% |
| F5 301B | 100% | 60% | THE OPEN GROUP OG0 092 | 50% | 50% |
| TEST PREP LSAT SECTION 2 READING COMPREHENSION | 100% | 60% | VERITAS VCS 323 | 50% | 50% |
| AVAYA 7004 | 100% | 50% | VERSA NETWORKS VNX100 | 50% | 50% |
| AVAYA 7591X | 100% | 50% | VMWARE 1V0 2120 | 50% | 50% |
| BCS ASTQB | 100% | 50% | VMWARE 2V0 2119 | 50% | 50% |
| CITRIX 1Y0 340 | 100% | 50% | VMWARE 2V0 2119D | 50% | 50% |
| CIW 1D0 520 | 100% | 50% | VMWARE 2V0 6120 | 50% | 50% |
| COMPTIA 220 1101 | 100% | 50% | VMWARE 2V0 6221 | 50% | 50% |
| COMPTIA CV1 003 | 100% | 50% | VMWARE 2V0 642 | 50% | 50% |
| CWNP PW0 071 | 100% | 50% | VMWARE 3V0 4220 | 50% | 50% |
| DELL DEA 5TT1 | 100% | 50% | VMWARE 5V0 23 20 | 50% | 50% |
| DELL DES 5121 | 100% | 50% | VMWARE 5V0 35 21 | 50% | 50% |
| ECCOUNCIL 312 49V8 | 100% | 50% | VMWARE 5V0 61 22 | 50% | 50% |

| Certification | Turbo-GPT3.5 | GPT3 | Certification | Turbo-GPT3.5 | GPT3 |
|---|---|---|---|---|---|
| ECCOUNCIL EC0 349 | 100% | 50% | VMWARE 5V0 6219 | 50% | 50% |
| EXIN EX0 002 | 100% | 50% | ZEND 200 710 | 50% | 50% |
| HITACHI HQT 4180 | 100% | 50% | GIAC GISF | 50% | 40% |
| ISTQB CTAL TM | 100% | 50% | HUAWEI H12 224 | 50% | 40% |
| MICROSOFT 70 537 | 100% | 50% | SANS SEC504 | 50% | 40% |
| MICROSOFT MB 600 | 100% | 50% | TEST PREP CFA LEVEL 3 | 50% | 40% |
| MICROSOFT MB2 712 | 100% | 50% | ISC CAP | 50% | 30% |
| MICROSOFT PL 100 | 100% | 50% | AIWMI CCRA | 50% | 25% |
| SAP C HANATEC 17 | 100% | 50% | ALCATEL LUCENT 4A0 M02 | 50% | 25% |
| SAP E ACTCLD 21 | 100% | 50% | APPIAN ACD200 | 50% | 25% |
| SAS INSTITUTE A00 250 | 100% | 50% | ARUBA ACMP 64 | 50% | 25% |
| SIX SIGMA LSSWB | 100% | 50% | AVAYA 3301 | 50% | 25% |
| SNIA S10 210 | 100% | 50% | AVAYA 3304 | 50% | 25% |
| SOA S9001 | 100% | 50% | AVAYA 6209 | 50% | 25% |
| TEST PREP CBEST SECTION 2 READING | 100% | 50% | AVAYA 71200X | 50% | 25% |
| VERITAS VCS 413 | 100% | 50% | AVAYA 71201X | 50% | 25% |
| VMWARE 2V0 5121 | 100% | 50% | AVAYA 7141X | 50% | 25% |
| ECCOUNCIL 312 49V10 | 100% | 40% | AVAYA 7304 | 50% | 25% |
| EXIN ITILF | 100% | 40% | BACB BCBA | 50% | 25% |
| LINUX FOUNDATION LFCS | 100% | 40% | CITRIX 1Y0 341 | 50% | 25% |
| MICROSOFT AZ 720 | 100% | 33% | CITRIX 1Y0 401 | 50% | 25% |
| MICROSOFT PL 400 | 100% | 33% | COMPTIA DA0 001 | 50% | 25% |
| VMWARE 1V0 642 | 100% | 33% | DATABRICKS CERTIFIED ASSOCIATE DEVELOPER FOR APACHE SPARK | 50% | 25% |
| ASQ CQA | 100% | 25% | DELL DCPPE 200 | 50% | 25% |
| CIW 1D0 610 | 100% | 25% | DELL DES 1221 | 50% | 25% |
| DELL E05 001 | 100% | 25% | DELL DES 1241 | 50% | 25% |
| HP HPE0 J74 | 100% | 25% | DELL DES 9131 | 50% | 25% |
| LPI 101 400 | 100% | 25% | DELL E20 020 | 50% | 25% |
| MICROSOFT 70 535 | 100% | 25% | DELL E20 575 | 50% | 25% |
| MICROSOFT 70 767 | 100% | 25% | DELL E20 655 | 50% | 25% |
| MICROSOFT 70 776 | 100% | 25% | DELL E20 920 | 50% | 25% |
| MICROSOFT 70 778 | 100% | 25% | ECCOUNCIL 312 76 | 50% | 25% |
| MICROSOFT MB 340 | 100% | 25% | FORTINET FORTISANDBOX | 50% | 25% |
| MICROSOFT MS 301 | 100% | 25% | FORTINET NSE5 FMG 6 4 | 50% | 25% |
| NETAPP NS0 171 | 100% | 25% | FORTINET NSE6 FAC 6 1 | 50% | 25% |
| SALESFORCE CERTIFIED SHARING AND VISIBILITY ARCHITECT | 100% | 25% | FORTINET NSE8 810 | 50% | 25% |

| Certification | Turbo-GPT3.5 | GPT3 | Certification | Turbo-GPT3.5 | GPT3 |
|---|---|---|---|---|---|
| SAS INSTITUTE A00 212 | 100% | 25% | FORTINET NSE8 811 | 50% | 25% |
| SAS INSTITUTE A00 240 | 100% | 25% | GENESYS CIC 101 01 | 50% | 25% |
| DELL E20 357 | 100% | 0% | GOOGLE PROFESSIONAL CLOUD DATABASE ENGINEER | 50% | 25% |
| MICROSOFT 70 331 | 100% | 0% | GUIDANCE SOFTWARE GD0 100 | 50% | 25% |
| MICROSOFT 70 466 | 100% | 0% | HP HPE0 J79 | 50% | 25% |
| MICROSOFT 70 705 | 100% | 0% | HP HPE0 S22 | 50% | 25% |
| MICROSOFT 98 367 | 100% | 0% | HP HPE6 A42 | 50% | 25% |
| MICROSOFT AZ 202 | 100% | 0% | ISACA CCAK | 50% | 25% |
| MICROSOFT MB 910 | 100% | 0% | ISTQB CTFL 2018 | 50% | 25% |
| MICROSOFT MB6 894 | 100% | 0% | LPI 202 450 | 50% | 25% |
| SOLARWINDS SCP 500 | 100% | 0% | MAGENTO M70 201 | 50% | 25% |
| VMWARE 2V0 2120 | 100% | 0% | META FACEBOOK 100 101 | 50% | 25% |
| VMWARE 2V0 631 | 100% | 0% | MICROSOFT 70 243 | 50% | 25% |
| ISC CCSP | 90% | 80% | MICROSOFT 70 346 | 50% | 25% |
| ALCATEL LUCENT 4A0 100 | 90% | 70% | MICROSOFT 70 765 | 50% | 25% |
| APICS CSCP | 90% | 70% | MICROSOFT 70 774 | 50% | 25% |
| ASIS ASIS CPP | 90% | 70% | MICROSOFT AZ 100 | 50% | 25% |
| COMPTIA SY0 401 | 90% | 70% | MICROSOFT MB 310 | 50% | 25% |
| MICROSOFT MS 100 | 90% | 70% | NETAPP NS0 155 | 50% | 25% |
| PEOPLECERT 58 | 90% | 70% | NETAPP NS0 162 | 50% | 25% |
| TEST PREP NCLEX RN | 90% | 70% | NETAPP NS0 175 | 50% | 25% |
| ACSM 010 111 | 90% | 60% | NETAPP NS0 191 | 50% | 25% |
| ECCOUNCIL 712 50 | 90% | 60% | NFPA CFPS | 50% | 25% |
| EXIN EX0 001 | 90% | 60% | NUTANIX NCA | 50% | 25% |
| TEST PREP MCAT TEST | 90% | 60% | PALO ALTO NETWORKS PCCSE | 50% | 25% |
| TEST PREP USMLE | 90% | 60% | PEGASYSTEMS PEGACPBA73V1 | 50% | 25% |
| ISACA CRISC | 90% | 50% | SALESFORCE CERTIFIED IDENTITY AND ACCESS MANAGEMENT DESIGNER | 50% | 25% |
| SIX SIGMA ICBB | 90% | 50% | SALESFORCE CERTIFIED INDUSTRIES CPQ DEVELOPER | 50% | 25% |
| SIX SIGMA LSSBB | 90% | 50% | SALESFORCE FIELD SERVICE CONSULTANT | 50% | 25% |
| ECCOUNCIL 312 49 | 90% | 20% | SERVICENOW CAS PA | 50% | 25% |
| COMPTIA N10 008 | 89% | 89% | SERVICENOW CIS APM | 50% | 25% |
| MICROSOFT MD 100 | 89% | 78% | SERVICENOW CIS DISCOVERY | 50% | 25% |
| COMPTIA N10 006 | 89% | 67% | SERVICENOW CIS SIR | 50% | 25% |

| Certification | Turbo-GPT3.5 | GPT3 | Certification | Turbo-GPT3.5 | GPT3 |
|---|---|---|---|---|---|
| MICROSOFT AZ 204 | 88% | 88% | TEST PREP ACLS | 50% | 25% |
| MICROSOFT MS 500 | 88% | 88% | TEST PREP GED SECTION 1 SOCIAL STUDIES | 50% | 25% |
| MICROSOFT MS 101 | 88% | 63% | TEST PREP HESI A2 | 50% | 25% |
| COMPTIA CS0 001 | 86% | 57% | TEST PREP RPFT | 50% | 25% |
| TEST PREP CGFM | 80% | 100% | VERITAS VCS 277 | 50% | 25% |
| ECCOUNCIL 312 50V10 | 80% | 90% | VMWARE 1V0 601 | 50% | 25% |
| SIX SIGMA LSSMBB | 80% | 90% | VMWARE 1V0 605 | 50% | 25% |
| AHIMA RHIA | 80% | 80% | VMWARE 2V0 2119 PSE | 50% | 25% |
| AMAZON AWS CERTIFIED MACHINE LEARNING SPECIALTY | 80% | 80% | VMWARE 2V0 3119 | 50% | 25% |
| AMAZON AWS CERTIFIED SYSOPS ADMINISTRATOR ASSOCIATE | 80% | 80% | VMWARE 2V0 4119 | 50% | 25% |
| CHECKPOINT 156 215 81 | 80% | 80% | VMWARE 2V0 72 22 | 50% | 25% |
| COMPTIA FC0 U61 | 80% | 80% | VMWARE 2V0 751 | 50% | 25% |
| MICROSOFT 70 697 | 80% | 80% | DELL E20 593 | 50% | 20% |
| MICROSOFT AZ 500 | 80% | 80% | SALESFORCE CERTIFIED PLATFORM DEVELOPER II | 50% | 20% |
| SCRUM PSM I | 80% | 80% | AMAZON AWS CERTIFIED BIG DATA SPECIALTY | 50% | 0% |
| CHECKPOINT 156 21577 | 80% | 70% | AVAYA 3002 | 50% | 0% |
| TEST PREP GMAT SECTION 3 | 80% | 70% | AVAYA 7130X | 50% | 0% |
| TEST PREP GMAT SECTION 3 VERBAL ABILITY | 80% | 70% | AVAYA 7492X | 50% | 0% |
| THE OPEN GROUP OG0 091 | 80% | 70% | AVAYA 76940X | 50% | 0% |
| COMPTIA 220 901 | 80% | 60% | CHECKPOINT 156 585 | 50% | 0% |
| COMPTIA SK0 005 | 80% | 60% | DELL DES 1D11 | 50% | 0% |
| GIAC GCIH | 80% | 60% | FORTINET NSE4 FGT 70 | 50% | 0% |
| GOOGLE PROFESSIONAL CLOUD ARCHITECT | 80% | 60% | FORTINET NSE6 FML 538 | 50% | 0% |
| HASHICORP TERRAFORM ASSOCIATE | 80% | 60% | FORTINET NSE6 FWB 61 | 50% | 0% |
| PMI PMI SP | 80% | 60% | FORTINET NSE6 FWF 64 | 50% | 0% |
| SALESFORCE ADM 211 | 80% | 60% | HITACHI HCE 3700 | 50% | 0% |
| TEST PREP CDL | 80% | 60% | HP HPE0 J68 | 50% | 0% |
| AMAZON AWS CERTIFIED SECURITY SPECIALTY | 80% | 50% | HP HPE0 J75 | 50% | 0% |
| COMPTIA CV0 001 | 80% | 50% | HP HPE0 Y53 | 50% | 0% |

| Certification | Turbo-GPT3.5 | GPT3 | Certification | Turbo-GPT3.5 | GPT3 |
|---|---|---|---|---|---|
| COMPTIA PK0 004 | 80% | 50% | MICROSOFT 70 341 | 50% | 0% |
| COMPTIA TK0 201 | 80% | 50% | MICROSOFT AZ 102 | 50% | 0% |
| GAQM APM 001 | 80% | 50% | MICROSOFT MB2 714 | 50% | 0% |
| GAQM CLSSBB | 80% | 50% | MICROSOFT MB2 877 | 50% | 0% |
| GOOGLE ADWORDS FUNDAMENTALS | 80% | 50% | NOKIA BL0 100 | 50% | 0% |
| GOOGLE INDIVIDUAL QUALIFICATION | 80% | 50% | PEGASYSTEMS PEGACPMC74V1 | 50% | 0% |
| SALESFORCE CRT 450 | 80% | 50% | PEGASYSTEMS PEGAPCSA80V1 2019 | 50% | 0% |
| TEST PREP ACT TEST | 80% | 50% | PEGASYSTEMS PEGAPCSSA87V1 | 50% | 0% |
| BLUE COAT BCCPA | 80% | 40% | SAP C SAC 2221 | 50% | 0% |
| ISC CISSP ISSMP | 80% | 40% | SAP C TS450 2020 | 50% | 0% |
| MICROSOFT DP 201 | 80% | 40% | SAP C TSCM62 67 | 50% | 0% |
| PMI CAPM | 80% | 40% | SERVICENOW CIS CSM | 50% | 0% |
| TEST PREP MCQS | 80% | 40% | SNOWFLAKE SNOWPRO ADVANCED ARCHITECT | 50% | 0% |
| TEST PREP SAT SECTION 1 CRITICAL READING | 80% | 40% | SOA S9002 | 50% | 0% |
| AMAZON AWS CERTIFIED CLOUD PRACTITIONER | 80% | 30% | VEEAM VMCE V9 | 50% | 0% |
| ALCATEL LUCENT 4A0 103 | 80% | 20% | VMWARE 1V0 701 | 50% | 0% |
| BLOCKCHAIN CBSA | 80% | 20% | VMWARE 2V0 731 | 50% | 0% |
| GIAC GSEC | 80% | 20% | VMWARE 2VB 601 | 50% | 0% |
| HUAWEI H12 221 | 80% | 20% | VMWARE 5V0 3219 | 50% | 0% |
| ISC CISSP ISSEP | 80% | 20% | MICROSOFT DP 200 | 43% | 43% |
| MICROSOFT 70 741 | 80% | 20% | SYMANTEC 250 513 | 40% | 70% |
| MICROSOFT 70 761 | 80% | 20% | ABA CTFA | 40% | 60% |
| MICROSOFT MD 101 | 78% | 67% | ALCATEL LUCENT 4A0 101 | 40% | 60% |
| ASQ CQE | 75% | 100% | BACB BCABA | 40% | 60% |
| ENGLISH TEST PREPARATION TOEFL SENTENCE CORRECTION | 75% | 100% | ECCOUNCIL 412 79V8 | 40% | 60% |
| FORTINET NSE5 FAZ 60 | 75% | 100% | MICROSOFT 70 744 | 40% | 60% |
| GOOGLE ASSOCIATE ANDROID DEVELOPER | 75% | 100% | SALESFORCE CERTIFIED ADVANCED ADMINISTRATOR | 40% | 60% |
| HP HPE2 E71 | 75% | 100% | SALESFORCE CERTIFIED SALES CLOUD CONSULTANT | 40% | 60% |
| IIBA ECBA | 75% | 100% | TEST PREP CFA LEVEL 2 | 40% | 60% |
| NETAPP NS0 145 | 75% | 100% | TEST PREP NET | 40% | 60% |

| Certification | Turbo-GPT3.5 | GPT3 | Certification | Turbo-GPT3.5 | GPT3 |
|---|---|---|---|---|---|
| PALO ALTO NETWORKS PSE SASE | 75% | 100% | AMAZON AWS SYSOPS | 40% | 40% |
| SALESFORCE CERTIFIED MARKETING CLOUD ADMINISTRATOR | 75% | 100% | CHECKPOINT 156 31580 | 40% | 40% |
| VMWARE 2V0 7121 | 75% | 100% | DELL E20 260 | 40% | 40% |
| A10 NETWORKS A10 CERTIFIED PROFESSIONAL SYSTEM ADMINISTRATION 4 | 75% | 75% | HP HPE6 A29 | 40% | 40% |
| AAFM INDIA CWM LEVEL 1 | 75% | 75% | PMI PFMP | 40% | 40% |
| AVAYA 7003 | 75% | 75% | SAS INSTITUTE A00 211 | 40% | 40% |
| AVAYA 7392X | 75% | 75% | TEST PREP PTCE | 40% | 40% |
| BLOCKCHAIN CBBF | 75% | 75% | TEST PREP SAT TEST | 40% | 40% |
| CHECKPOINT 156 315 81 | 75% | 75% | TEST PREP TCLEOSE | 40% | 40% |
| CITRIX 1Y0 231 | 75% | 75% | VMWARE 2V0 621 | 40% | 40% |
| CIW 1D0 541 | 75% | 75% | VMWARE 2V0 621D | 40% | 40% |
| COMPTIA CLO 002 | 75% | 75% | APICS CLTD | 40% | 20% |
| COMPTIA XK0 004 | 75% | 75% | BLUE COAT BCCPP | 40% | 20% |
| CROWDSTRIKE CCFA | 75% | 75% | F5 101 | 40% | 20% |
| CWNP CWAP 402 | 75% | 75% | FINRA SERIES 7 | 40% | 20% |
| CWNP CWNA 107 | 75% | 75% | GIAC GSSP JAVA | 40% | 20% |
| CWNP CWNA 108 | 75% | 75% | HRCI GPHR | 40% | 20% |
| DELL DEA 2TT3 | 75% | 75% | IAPP CIPM | 40% | 20% |
| DELL E10 002 | 75% | 75% | ISACA CDPSE | 40% | 20% |
| ECCOUNCIL 312 85 | 75% | 75% | SALESFORCE CERTIFIED MARKETING CLOUD EMAIL SPECIALIST | 40% | 20% |
| ENGLISH TEST PREPARATION TOEFL SENTENCE COMPLETION | 75% | 75% | TEST PREP PCAT | 40% | 20% |
| EXIN ISFS | 75% | 75% | CHECKPOINT 156 91577 | 40% | 10% |
| GENESYS GCP GC IMP | 75% | 75% | DELL E20 555 | 40% | 0% |
| GENESYS GCP GCX | 75% | 75% | RIVERBED 101 01 | 40% | 0% |
| GOOGLE PROFESSIONAL CLOUD DEVOPS ENGINEER | 75% | 75% | MICROSOFT MB 210 | 38% | 50% |
| GOOGLE PROFESSIONAL CLOUD NETWORK ENGINEER | 75% | 75% | DELL DES 1121 | 33% | 67% |
| GOOGLE SHOPPING ADVERTISING | 75% | 75% | HP HP0 Y47 | 33% | 67% |

| Certification | Turbo-GPT3.5 | GPT3 | Certification | Turbo-GPT3.5 | GPT3 |
|---|---|---|---|---|---|
| HP HPE0 S46 | 75% | 75% | HP HPE2 T37 | 33% | 67% |
| HP HPE6 A70 | 75% | 75% | MICROSOFT 70 742 | 33% | 67% |
| HP HPE6 A73 | 75% | 75% | MICROSOFT MS 740 | 33% | 67% |
| MAGENTO M70 301 | 75% | 75% | MICROSOFT SC 200 | 33% | 67% |
| MICROSOFT 70 463 | 75% | 75% | CHECKPOINT 156 31577 | 33% | 44% |
| MICROSOFT 70 685 | 75% | 75% | COMPTIA LX0 103 | 33% | 33% |
| MICROSOFT 70 773 | 75% | 75% | CYBERARK PAM CDE RECERT | 33% | 33% |
| MICROSOFT 74 409 | 75% | 75% | DELL DES 6332 | 33% | 33% |
| MICROSOFT DP 300 | 75% | 75% | DELL DES DD33 | 33% | 33% |
| MICROSOFT MB 230 | 75% | 75% | HP HP2 B149 | 33% | 33% |
| MICROSOFT MB2 708 | 75% | 75% | HP HPE6 A72 | 33% | 33% |
| MICROSOFT SC 300 | 75% | 75% | MICROSOFT 70 462 | 33% | 33% |
| NADCA ASCS | 75% | 75% | MICROSOFT 70 480 | 33% | 33% |
| NETAPP NS0 180 | 75% | 75% | MICROSOFT 70 496 | 33% | 33% |
| NETAPP NS0 194 | 75% | 75% | MICROSOFT DP 203 | 33% | 33% |
| NOVELL 050 733 | 75% | 75% | MICROSOFT MB 220 | 33% | 33% |
| PALO ALTO NETWORKS PSE STRATA | 75% | 75% | MICROSOFT MS 600 | 33% | 33% |
| SALESFORCE CERTIFIED BUSINESS ANALYST | 75% | 75% | MICROSOFT PL 900 | 33% | 33% |
| SAP C TAW12 750 | 75% | 75% | PEGASYSTEMS PEGACSA72V1 | 33% | 33% |
| SCALED AGILE SA | 75% | 75% | SALESFORCE DEV 401 | 33% | 33% |
| SIX SIGMA ICGB | 75% | 75% | VMWARE 2V0 602 | 33% | 33% |
| SIX SIGMA LSSGB | 75% | 75% | VMWARE 3V0 624 | 33% | 33% |
| SYMANTEC 250 430 | 75% | 75% | DELL E20 307 | 33% | 0% |
| TEST PREP TEAS SECTION 2 SENTENCE CORRECTION | 75% | 75% | FORTINET NSE5 FAZ 62 | 33% | 0% |
| VMWARE 1V0 603 | 75% | 75% | HP HPE0 S58 | 33% | 0% |
| VMWARE 2V0 6119 | 75% | 75% | MICROSOFT 70 473 | 33% | 0% |
| VMWARE 2V0 620 | 75% | 75% | MICROSOFT 70 695 | 33% | 0% |
| WORLDATWORK T1 GR1 | 75% | 75% | MICROSOFT 70 713 | 33% | 0% |
| AMAZON AWS CERTIFIED ADVANCED NETWORKING SPECIALTY ANS C01 | 75% | 50% | MICROSOFT 98 369 | 33% | 0% |
| ANDROIDATC AND 403 | 75% | 50% | PEGASYSTEMS PEGACRSA80V1 | 33% | 0% |
| APPLE MAC 16A | 75% | 50% | HUAWEI H13 629 | 30% | 40% |
| AVAYA 3107 | 75% | 50% | ISACA CGEIT | 30% | 30% |
| AVAYA 7120X | 75% | 50% | GIAC GPEN | 30% | 10% |

| Certification | Turbo-GPT3.5 | GPT3 | Certification | Turbo-GPT3.5 | GPT3 |
|---|---|---|---|---|---|
| AVAYA 7593X | 75% | 50% | AVAYA 6211 | 25% | 75% |
| BCS TM12 | 75% | 50% | AVAYA 7230X | 25% | 75% |
| BLOCKCHAIN CBDE | 75% | 50% | DELL E20 893 | 25% | 75% |
| CHECKPOINT 156 110 | 75% | 50% | GENESYS GCP GC REP | 25% | 75% |
| CIW 1D0 571 | 75% | 50% | HP HP2 B148 | 25% | 75% |
| COMPTIA LX0 104 | 75% | 50% | HP HPE0 S52 | 25% | 75% |
| ECCOUNCIL 212 89 | 75% | 50% | HP HPE6 A49 | 25% | 75% |
| EXIN EX0 105 | 75% | 50% | HUAWEI H19 301 | 25% | 75% |
| EXIN PR2P | 75% | 50% | CHECKPOINT 156 91580 | 25% | 50% |
| FORTINET NSE4 FGT 62 | 75% | 50% | CITRIX 1Y0 204 | 25% | 50% |
| FORTINET NSE5 FMG 70 | 75% | 50% | CITRIX 1Y0 311 | 25% | 50% |
| FORTINET NSE6 | 75% | 50% | CIW 1D0 621 | 25% | 50% |
| FORTINET NSE6 FWB 560 | 75% | 50% | DELL DES 1721 | 25% | 50% |
| FORTINET NSE7 EFW 70 | 75% | 50% | DELL DES 1B31 | 25% | 50% |
| FORTINET NSE7 PBC 64 | 75% | 50% | DELL DNDNS 200 | 25% | 50% |
| GIAC GCPM | 75% | 50% | ENGLISH TEST PREPARATION TOEFL READING COMPREHENSION | 25% | 50% |
| GOOGLE MOBILE ADVERTISING | 75% | 50% | EXIN ISMP | 25% | 50% |
| GOOGLE PROFESSIONAL COLLABORATION ENGINEER | 75% | 50% | FORTINET NSE7 | 25% | 50% |
| GOOGLE VIDEO ADVERTISING | 75% | 50% | GENESYS GE0 806 | 25% | 50% |
| HP HPE0 S37 | 75% | 50% | HP HPE2 E72 | 25% | 50% |
| HP HPE2 E69 | 75% | 50% | HP HPE6 A07 | 25% | 50% |
| HRCI PHR | 75% | 50% | INFOR IOS 252 | 25% | 50% |
| INFOR M3 123 | 75% | 50% | ISQI CTAL TA | 25% | 50% |
| ISTQB ATM | 75% | 50% | MCAFEE MA0 101 | 25% | 50% |
| ISTQB ATTA | 75% | 50% | MICROSOFT 74 343 | 25% | 50% |
| LPI 010 150 | 75% | 50% | MICROSOFT MB 240 | 25% | 50% |
| MICROSOFT 70 483 | 75% | 50% | NETAPP NS0 502 | 25% | 50% |
| MICROSOFT 70 486 | 75% | 50% | SALESFORCE CERTIFIED HEROKU ARCHITECTURE DESIGNER | 25% | 50% |
| MICROSOFT 70 735 | 75% | 50% | SALESFORCE CERTIFIED SERVICE CLOUD CONSULTANT | 25% | 50% |
| MICROSOFT 70 768 | 75% | 50% | SERVICENOW CIS EM | 25% | 50% |